\documentclass[letterpaper, 10 pt, conference]{ieeeconf}  

\IEEEoverridecommandlockouts                              
 
\usepackage{graphicx}
\usepackage{times} 
\usepackage{amsmath} 
\usepackage{amssymb}  
\usepackage[normalem]{ulem} 

\usepackage[usenames]{xcolor}

\title{
Grounding Perception: A Developmental Approach to Sensorimotor Contingencies
}

\author{Alban Laflaqui\`{e}re$^*$, Nikolas Hemion, Micha\"{e}l Garcia Ortiz, Jean-Christophe Baillie
\thanks{AI Lab, Aldebaran, 43 rue du Colonel Avia, 75015 Paris, France 
        {\tt\small \{alaflaquiere$^*$, nhemion, mgarciaortiz, jcbaillie\}@aldebaran.com}}%
\thanks{$^*$ corresponding author }
}

\begin{document}

\maketitle 
\thispagestyle{empty} 
\pagestyle{empty} 
 
\begin{abstract} 
Sensorimotor contingency theory offers a promising account of the nature of perception, a topic rarely addressed in the robotics community.
We propose a developmental framework to address the problem of the autonomous acquisition of sensorimotor contingencies by a naive robot. While exploring the world, the robot internally encodes contingencies as predictive models that capture the structure they imply in its sensorimotor experience.
Three preliminary applications are presented to illustrate our approach to the acquisition of perceptive abilities: discovering the environment, discovering objects, and discovering a visual field.
\end{abstract} 
 
\section{Introduction}
\label{sec:1}

Perception in robotics is traditionally addressed by the hand design of sophisticated feature extraction and classification algorithms, 
%
%
which leads to impressive results in a range of complex tasks, such as object recognition \cite{Andreopoulos2013827} or grasping and manipulation \cite{darpaARM}.
However, the vast majority of those algorithms have been developed to solve specialized tasks, and turn out to be inefficient or completely useless when  faced with unexpected situations. Moreover, they are typically fine-tuned for
specific
sensory and motor apparatus, and not easily adapted when
changed or broken.
Although such systems are progressively augmented to handle more varied situations, the 
fundamental problem
remains:
perceptual concepts that are defined by designers, instead of being
constructed by robots and grounded in their experience, are inflexible, and thus cannot engender genuine autonomy \cite{Ziemke}.

We consider the focus of roboticists on the question of \textit{``how to perceive"}, instead of \textit{``what is perception"}, to be a fundamental reason for these shortcomings. The nature of perception
is indirectly defined
through the development of feature extraction algorithms, to which roboticists have projected 
their
intuition on the matter (inspired, to some degree, by biology). Yet, these intuitions might be misleading, especially for robotic systems that prove to be notably different from biological ones.
We believe that answering this second question -- \textit{``what is perception"} -- is a preliminary step to overcome current limitations in robotics, and endow artificial systems with genuine perception. It requires finding a middle ground between the traditional hand-coded approach, in which human intuitions on perception are implemented in a rigid fashion, 
and 
behaviorist approaches,
in which the robot
learns
to solve a task without addressing the question of how it perceives the world \cite{Doncieux2014, Kober23082013}.

Coming from outside the realm of robotics, the seminal paper \cite{o2001sensorimotor} introduced sensorimotor contingencies theory (SMCT) as a groundbreaking account for the nature of perception. 
It proposes a definition of perception in an agent as \textit{``the mastering of sensorimotor contingencies''} without requiring a priori knowledge about the external environment. 
Although promising, its dissemination inside the robotics community has been slow for two reasons. First, it requires a 
paradigm shift from traditional approaches. Second, the theory still lacks a clear formalization, leaving its implementation on robots an unsolved problem.
Some computational models inspired by SMCT have already been described to address perceptive domains such as space \cite{terekhov2013space,laflaquiere2012non}, color \cite{philipona2006color,witzel2015determines}, and body schema \cite{laflaquiere2013learning}, but no mechanism has been proposed to explain how they could arise in a robot.

We propose to address the question of sensorimotor contingencies acquisition from a developmental point of view.
Studies on animals as well as humans suggest that perception is acquired on an epigenetic timescale.
%
%
Experiments show that this phenomenon not only occurs in infants, but also in adults \cite{bach2003sensory}, and that even previously acquired perceptive skills may be adapted or altered \cite{herwig2014predicting, linden1999myth}.
Based on these biological considerations, our approach to perception moves away from traditional robotics to fit into the developmental robotics framework \cite{cangelosi2015developmental}. It proposes that the robot, as an embodied agent, should explore the world and incrementally build its own way to perceive and interact with it. This requires the definition of a minimal set of mechanisms and constraints that would allow such development. Those mechanisms should be generic enough to adapt to diverse forms of perception encountered by the agent (e.g. different modalities) and varying levels of abstraction. 
To date, no clear formalism for those mechanisms has arisen in the developmental robotics community.

%
%
We propose predictive modeling \cite{wolpert1995internal, friston2006free} as such a computational mechanism to learn sensorimotor contingencies, and thus acquire perceptive skills.
In the context of SMCT, predictive models can be autonomously estimated by the agent to capture structure in the way motor commands actively transform sensory inputs, namely sensorimotor contingencies.
Predictive modeling allows the incremental acquisition of skills required in developmental robotics, while providing a computational implementation of the concept of sensorimotor contingencies.


Our current implementation of the formalism proposed in this paper uses a method to cluster state transition graphs, to discover densely connected subgraphs. Note that similar methods have already been proposed by others, for example in navigation tasks for the segmentation of location data into rooms \cite{zivkovic2006hierarchical}, or for sub-goal discovery in hierarchical reinforcement learning (e.g. \cite{mannor2004dynamic,csimcsek2005identifying}). The contribution of this work is however not on the algorithmic, but on the computational level \cite{marr1976understanding}: we aim at shedding light on the question of how sensorimotor contingencies can be discovered and captured autonomously by a naive agent.


The current state of our formalism is presented in more detail in Sec.~\ref{sec:2}. We propose preliminary results illustrating the approach in Sec.~\ref{sec:preliminary applications}. We will discuss the benefits and limitations of the approach in Sec.~\ref{sec:ccl}, and propose possible future developments in order to reach our objective: developing a unified framework for autonomous sensorimotor contingency acquisition.


\section{A developmental framework for SMCT}
\label{sec:2}

The presentation of our developmental approach of sensorimotor contingencies is divided into three parts. First, we develop the idea that SMC define a perceptive ontology. Second, we introduce predictive modeling as a computational implementation of contingencies. Third, we describe an approach to allow a naive agent to build its own predictive models, and interpret the world with which it interacts.

\subsection{SMC as a perceptive ontology for naive agents}
\label{sec:2-1}

An agent, etymologically ``that which acts", is characterized primarily by its ability to generate motor commands,
which
can be further influenced by taking into account information from the environment (i.e. sensory inputs), and adapting generated actions (i.e. motor outputs) accordingly.
When this adaptation is direct and uncircumventable, we say that the agent is \textit{reacting} to the environment, but not \textit{perceiving} it. 
This may be observed
in systems such as
Braitenberg's vehicles \cite{braitenberg1986vehicles}, and robots controlled by artificially evolved neural networks \cite{brooks1989robot}.
However, when the choice of action is indirectly guided by the predicted outcome of actions, we claim that the agent \textit{perceives} the environment.
We define \textit{perception} as \textit{the ability to categorize sensory inputs based on the way they can be actively transformed by the agent}.

The relevance of such a definition of perception is particularly evident when considering the problem of a naive agent that needs to acquire perceptive abilities. Without a priori knowledge, labels, and predefined sensory categories, a naive agent cannot interpret any sensory input. 
Moreover, sensory pre-processing \cite{bengio2009learning, lowe2004distinctive} does not circumvent this problem; it merely generates a different input encoding (hopefully of lower dimension) that remains uninterpretable to the naive agent.
In order to adapt its actions, the agent must instead learn
how sensory inputs can be actively transformed by its motor commands. 
Then the agent can process any sensory input, by predicting these transformations, and potentially selecting the best action to reach a goal state (i.e. a different, target sensory input). Thus, for the agent, a sensory input is not relevant by itself, but only through the potentiality of transformations it allows in the sensorimotor space.

One interesting particularity of this approach is that sensory inputs can be transformed in specific ways depending on the actual property of the environment they encode. This is what SMCT describes as \textit{contingencies}.
For instance, if the agent interacts with a line, the sensory input is invariant to a certain motor command: the one that translates the sensor along the line. On the other hand, if the agent interacts with a circle, this sensory input is invariant to a different motor command: the one that rotates the sensor around the center of the circle\footnote{Note that these invariances to motor commands only partially characterize the sensorimotor transformations related to lines and circles, and are used here as simple illustrative examples.}. As illustrated by these simple examples, properties of the environment are reflected in the possible sensory transformations. 
Contingencies can thus be used to categorize sensory inputs based on their associated transformations; they define a perceptive ontology.

The above interpretation of sensorimotor contingencies is compatible with the core aspects of SMCT, while formulating concepts in a form suitable for computational methods, as described in Sec.~\ref{sec:2-2}.

\subsection{Capturing SMC as predictive models}
\label{sec:2-2}


As discussed in Sec.~\ref{sec:2-1}, sensorimotor contingencies are regularities, in the ways motor commands transform sensory inputs, that reflect the actual properties of the environment with which the agent interacts.
In other words, they are highly predictable sensorimotor transformations.
For the agent, \textit{mastering} a contingency means knowing the corresponding sensorimotor transformation.

In robotics and neuroscience literature, the encoding of knowledge relating sensory to motor information in a system (artificial or biological brain) is referred to as \emph{internal modeling}
%
%
\cite{wolpert1995internal}.
The role of internal models is to allow the agent to predict future sensorimotor states without having to effectively engage in an interaction. Predictive models are often considered necessary for advanced cognitive functions like planning,
%
%
but we subscribe to the view put forward by the SMC theory that they also sustain the ability to perceive by capturing regularities in sensorimotor transformations.

Predictive modeling of the sensorimotor space using local specialized representations has been proposed in \cite{sutton2011horde}. Our approach is different in the sense that we want to capture the structure that underlies the predictability of the sensorimotor experience, namely sensorimotor contingencies. We believe that this difference  will grant the potential ability to tackle ambiguous sensorimotor situations.

It is often assumed, in robotics literature involving internal models, that the agent has full access to all information that is relevant to the internal model (i.e. they can be directly measured by its sensors). However, more generally, it has to be assumed that hidden factors can influence the interplay between sensor and motor states. As proposed in \cite{Philipona2003}, for a naive agent that has only access to its sensory and motor flows, this interplay can be characterized by a function:
%
\begin{equation}
S = \phi_{E}(M),
\label{eq:phi function}
\end{equation}
%
with $S$ the sensory state of the agent, $M$ its motor state, $E$ the unknown state of the environment and $\phi$ the unknown sensorimotor law defined by the constraints the world applies to the sensorimotor experience.
Obviously, $\phi$ is tremendously complex for any real-world scenario, and it would be intractable to try to directly capture it in a single internal model.
Two co-occurring mechanisms exist to reduce the difficulty of the problem. The sensorimotor data can be compacted using an unsupervised encoding scheme inspired by representation learning \cite{bengio2009learning}, which will reduce the dimensionality of the sensorimotor data. Also, the agent can try to estimate $\phi$ by parts, dividing the problem into simpler sub-functions that are easier to model in (relevant) subparts of the sensorimotor space.



Taking again the example of a line, the agent can discover one contingency without trying to capture the whole function $\phi$: a certain input in a sensory subspace is invariant to a certain motor variation in a motor subspace. The sensory subspace is the one related to the sensor interacting with the line (while the sensory space can be of higher dimension) and the motor subspace is the one related to the set of motors that can make this sensor move.
Having a model of this simple sub-function, the agent can predict a subpart of its sensorimotor experience.




\subsection{Learning predictive models for SMC}
\label{sec:2-3}

\begin{figure}[t]
\includegraphics[width=1\linewidth]{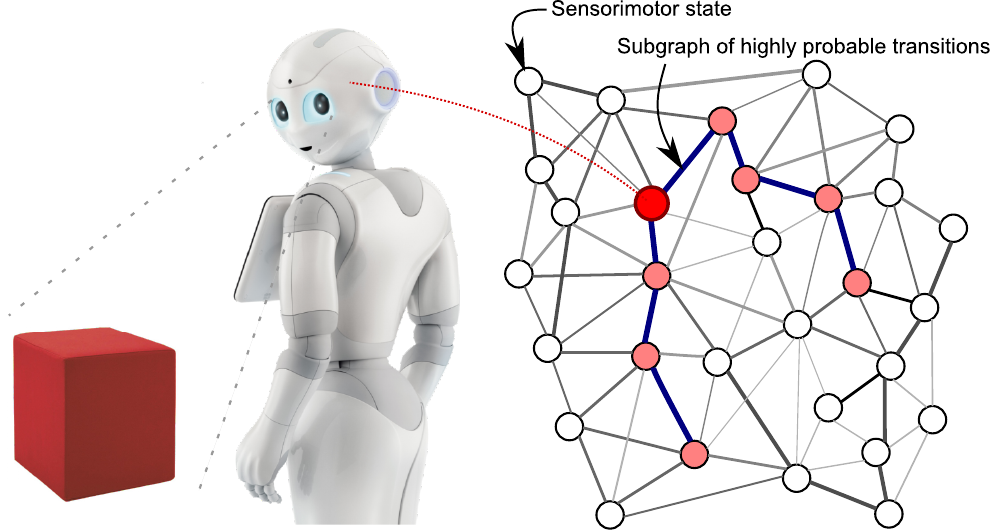}
\caption{
Schematic illustration. The agent's experience is a walk into the sensorimotor space. When experiencing a contingency, only a subgraph can be actively visited by the agent with a high probability .
}
\label{fig:contingency graph}
\end{figure}



From a developmental point of view, a naive agent has no a priori knowledge about any structure in its sensorimotor experience, and thus needs to learn sensorimotor contingencies. However, in doing so, it faces a dual problem: on the one hand, it needs to find a compact, efficient encoding of knowledge about the contingent transformation of sensorimotor states; on the other hand, it also has to discover by itself \emph{what} is predictable, and \emph{when} this is the case.

%
%
The former problem has long been studied in the machine learning and robotics literature, where many formalisms have been proposed to capture internal models (e.g. \cite{jordan1992forward, atkeson1997locally}).
Each method may exhibit certain benefits over others (for example: generalization capability, precision of learned estimates, or computational complexity), and typically the 
designer decides which method promises best performance for a given task and robot.
 
The focus of our work lies on the latter problem: how the agent can autonomously discover which sensorimotor sub-functions to model (the question of their optimal encoding should be addressed in future work).
From the perspective of the agent, this is a difficult task: it cannot simply learn everything it observes but has to decide by itself what parts of its interaction with the environment constitute a predictable pattern. Furthermore, as discussed above, the sensory outcome of its actions can depend on the current state of the environment. However it has no direct access to any information about this environmental state (for example, in the form of labels as in supervised learning), nor about the times when the environment changes (as in pre-segmented training data). In Sec.~\ref{sec:preliminary applications} we will describe examples to further illustrate these points and show how our method (described below) can be used to address them.





We assume that the sets of sensory and motor states are discrete and consider sensorimotor transitions between those states as building blocks for our internal models
(similar to state transition models in reinforcement learning \cite{sutton1998reinforcement}).
The experience of the agent can be seen as a graph in which each node corresponds to a sensorimotor state, and each edge to a sensorimotor transition between the two adjacent states.
The interaction of the agent with its environment corresponds to a walk through this graph.
Importantly, while the agent is engaged in an interaction corresponding to a sensorimotor contingency, only a subset of nodes can be visited (as illustrated in Fig.~\ref{fig:contingency graph}). While interacting with the contingency, transitions between nodes in this subgraph are very probable. On the other hand, transitions going out of the subgraph have a low probability, corresponding to the low probability of the environment suddenly changing its properties, and thus changing the sensorimotor contingency the agent was experiencing.
In order to learn a contingency, the agent therefore has to build a predictive model that captures the highly probable set of sensorimotor states and transitions that define the corresponding subgraph.
It can use this model to identify the contingency the next time it is encountered, and statistically predict what would be the sensorimotor outcome of its actions as a walk into this subgraph.
%

A very promising feature of the approach is that each identified subgraph can be promoted as a higher-level sensory input (or potentially a higher level motor command, depending on the kind of contingency captured). By identifying contingencies, the agent thus creates new sensorimotor spaces and related graphs (or extends existing ones) in which higher-level contingencies can be captured. 
This allows the learning of a hierarchical structure of contingencies. First, this is in line with the fact the natural world is hierarchical, structured and compositional. Second, it is highly relevant in terms of information processing, because the learned contingencies can be re-used and shared with multiple higher level contingencies, which is efficient and less redundant compared to a flat, shallow approach.
Third, an incremental construction of a representational hierarchy, such as this hierarchical construction of models for sensorimotor contingencies, is likely to be a key ingredient in any attempt to build a developmental cognitive system~\cite{guerin2013survey}.

Note that those predictive models can be estimated without assuming any specific policy while the agent explores the environment. Yet, some behaviors of the agent could limit the discovery of contingencies if their corresponding sensorimotor transitions are never experienced.
In the following applications of section \ref{sec:preliminary applications}, we consider random exploratory behaviors similar to motor babbling \cite{saegusa2009active}.

\begin{figure*}[t]
\includegraphics[width=1\linewidth]{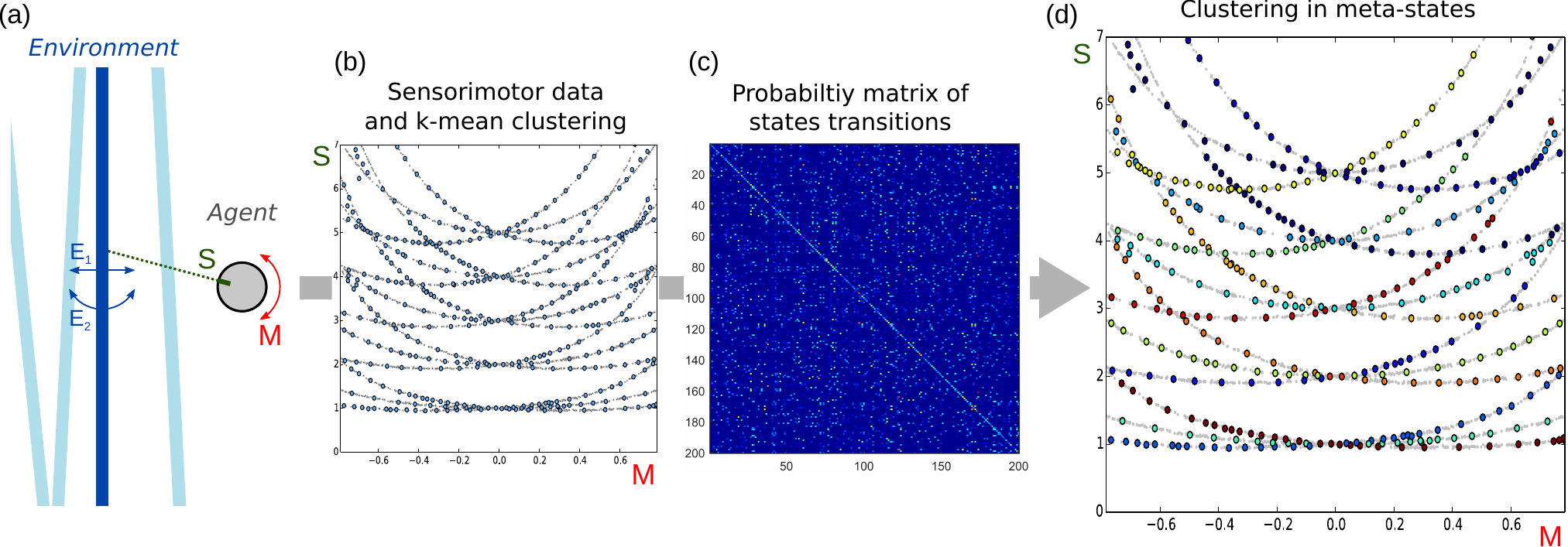}
\caption{
(a) Agent rotating and measuring its distance to a moving wall. (b) Clustering of the collected sensorimotor data using k-means
(each line in the sensorimotor space corresponds to a single position of the wall).
(c) Matrix $T$ of estimated probabilities of transitions between the sensorimotor clusters (only the first $200$ rows and columns are displayed for the sake of visualization). (d) Spectral clustering is applied to identify highly probable subgraphs in the matrix $T$. Each cluster is colored according to the subgraph it belongs to, which shows that each environmental state has been internally encoded by a corresponding subgraph.
}
\label{fig:discovering environment}
\end{figure*}

In the applications presented in Sec.~\ref{sec:preliminary applications}, predictive models are estimated based on the probability of sensorimotor transitions, formalized either in absolute form,
\begin{equation}
P(S_j,M_l|S_i,M_k),
\label{eq:transition probability}
\end{equation}
expressing the probability of observing sensory states $S_j$ in motor state $M_l$ after having observed $S_i$ in motor state $M_k$, or in differential form,
\begin{equation}
P( S_j | \Delta M_q , S_i ),
\label{eq:transition probability}
\end{equation}
where $\Delta M_q$ denotes a motor variation.
Those transitions probabilities are stored in a matrix $T$ that will be processed using spectral clustering \cite{ng2002spectral} in order to cluster the data into consistent subgraphs of sensorimotor transitions.\\

Many problems related to the acquisition of sensorimotor contingencies have been raised in this description of the approach. In next section, simple scenarios are designed to illustrate some of the difficulties faced by a naive agent, and how the presented approach is applied to extract contingencies from its sensorimotor experience.

\section{Preliminary applications}
\label{sec:preliminary applications}

Our approach is illustrated by three preliminary simulations that address the acquisition of different perceptive capacities, namely the perception of the environment, the perception of objects and the acquisition of a visual field. The objective is
to briefly illustrate how the approach applies to different problems. Simulations are purposely simple in order to facilitate the results analyses and to limit computational cost. Note that a more exhaustive description of each experiment is, or will be, available in separate papers for the interested reader.

\subsection{Discovering the environment}
\label{sec:discovering the environment}

In this first simulation, we address the question of \textit{how a naive agent can discover the concept of environment}. This notion is of course fundamental for the development of future perceptive abilities, but not trivial for a naive agent. Indeed, how can it discover that its internal sensorimotor experience is actually defined by both its own state as well as the state of an external entity, namely the \textit{environment}? 

To study the question, we introduce the simple agent-environment system of Fig.~\ref{fig:discovering environment}. The agent has one motor that controls the position of one sensor, sensible to the distance to a wall that constitutes the environment.
From an internal point of view, the agent's experience reduces to a manifold in its sensorimotor space, representing the sensorimotor print of the environment (see Fig.~\ref{fig:discovering environment}). 

If the environment is fixed, the agent has no reason to assume its existence, because the associated sensorimotor experience is completely predictable and controllable. However, if the state of the environment can change (as with, for example, the wall moving in Fig.~\ref{fig:discovering environment}), the agent can actively experience a different manifold in its sensorimotor space for each state of the environment.
It ensues that the agent's experience is not entirely controllable and predictable unless it assumes the existence of different subgraphs (contingencies) corresponding to different states of an external entity, the \textit{environment}.

Our objective is for the agent to build predictive models of those subgraphs from its sensorimotor experience of the world.
We propose the following algorithm to solve the problem:
\begin{itemize}
\item Randomly explore the motor space (motor babbling) while the environmental state may randomly change, and collect resulting sensorimotor states $(S_i,M_k)$ and their transitions $(S_i,M_k \rightarrow S_j,M_l)$.

\item Cluster the collected sensorimotor data $(S_i,M_k)$ into $K=430$ clusters $X_p$ using k-means clustering in order to discretize the sensorimotor space.

\item Evaluate the probabilities of transitions $(X_p \rightarrow X_q)$ based on the experienced transitions and store them in a transition matrix $T$ with entries $T_{pq} = P(X_p|X_q)$.

\item Cluster the states $X_p$ into subgraphs $Y_a$ of highly predictable co-occurring transitions using spectral clustering on matrix $T$ \cite{ng2002spectral}.

\end{itemize}
Spectral clustering finds a partitioning of the rows in $T$, corresponding to densely connected sub-graphs in the corresponding graph. This partitioning corresponds to candidate hypotheses of sensorimotor contingencies.
The outcome of the algorithm is a classification of sensorimotor states $(S_i,M_k)$:  each $(S_i,M_k)$ belongs to a cluster $X_p$ which itself belongs to a subgraph $Y_a$. A visualization of the final clustering for a set of $15$ different environmental states is presented in Fig.~\ref{fig:discovering environment}. Note how the different subgraphs $Y_a$ internally correspond to the different environmental states that can be observed from an external point of view. As a result, each sensorimotor state is correctly associated with the corresponding environmental state that generated it (internally represented by $Y_a$).

Finally, given a sensorimotor state, the agent can evaluate the current state of the environment -- or more precisely,  the currently active internal subgraph -- and predict the sensorimotor outcome of its motor commands. Moreover, when the environmental state changes, the resulting sensorimotor transition can be interpreted by the agent as a transition between subgraph that it cannot control with motor commands; this change is thus \textit{external}.

One important remark is that the final clustering into subgraphs is successful only if the probability of experiencing a environmental change is lower than the probability of active motor changes during the exploration phase. Otherwise, the probabilities of states transitions $(X_p \rightarrow X_q)$ 
would not be significantly different, whether the two sensorimotor states belong to a single environmental state or to two different environmental states,
and the spectral clustering would thus fail to capture structure in the data. Intuitively, such a limitation makes sense. It corresponds to the inability of a naive agent to build an internal representation of its own ability to act and of the environment when this last is constantly changing in a random way.
Finally, we can notice in Fig.~\ref{fig:discovering environment} that some ambiguous points could be classified in three different meta-states. Given the current algorithm, the agent statistically clusters each point in a single meta-state.
Some extensions of the algorithm are under development to determine that sensorimotor states can belong to multiple meta-states.

\subsection{Discovering objects} 
\label{sec:discovering objects} 

\begin{figure*}[t]
\includegraphics[width=1\linewidth]{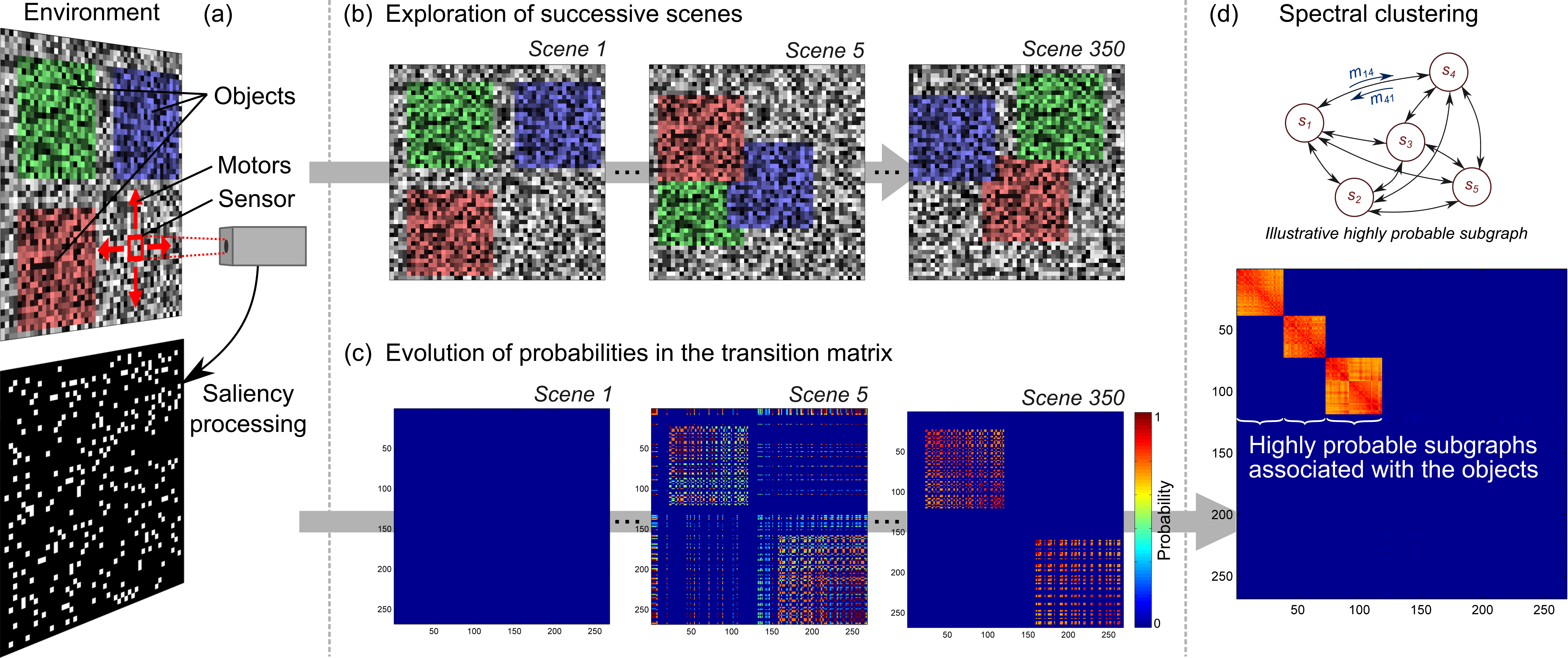}
\caption{
(a) Agent exploring its environment which includes three objects (colored for the sake of visualization). Only 'salient' sensory inputs are considered during exploration to reduce computational cost.
(b) Successive random scenes are created by moving the objects and change the environment with a $5\%$ probability.
(c) A transition matrix $T$ is created during the initial scene exploration. The probability of those sensorimotor transitions is then updated as other scenes are explored.
(d) Spectral clustering is applied to matrix $T$ to identify highly probable subgraphs and reorder its columns and rows accordingly. Three highly subgraphs corresponding to the three objects are correctly identified.
}
\label{fig:discovering objects}
\end{figure*}

In this second simulation, we address the question of \textit{how a naive agent can discover the concept of objects}. 
Compared to the concept of \textit{environments} introduced in \ref{sec:discovering the environment}, an object is a part of the environment that can be moved, or even be encountered in different environments. This property of independence from its surrounding is what makes it an interesting concept to discover for the agent. In previous simulation, the environment could have been made of different objects moving as a whole and the agent would have no way (or reason) to capture this fact.
This intuitive description can be translated into a sensorimotor definition: an \textit{object} 
is a consistent subgraph defined with differential sensorimotor transitions $(S_i,\Delta M_q \rightarrow S_j)$. This way, the object can be experienced from different starting motor states.

Our objective is for an naive agent to build internal representations of objects as such subgraphs.
In order to illustrate the approach, let us consider the agent-environment system of Fig.~\ref{fig:discovering objects}. The agent lives in a 2D world where the environment is made of square elements with random properties represented by a single value. A visual analogy of those properties could for instance be their luminance and/or color. Three objects are defined by randomly creating three sets of $20 \times 20$ elements (objects are made square for the sake of simplicity). Those objects are placed in the environment, overlapping already present elements. During the agent's exploration of the world, the environment can randomly change (redrawing the value of every element) and the objects can be placed at random locations, thus defining different "scenes" the agent can explore.
The agent is equipped with a sensor that captures the values of $3 \times 3$ neighbors elements in order to generate a 9D sensory input (see Fig.~\ref{fig:discovering objects}). It is also equipped with two motors that translate the sensor horizontally and vertically with steps corresponding to multiples of an element's length. 

We propose the following algorithm the let the agent capture objects as subgraphs in the sensorimotor space:
%
\begin{itemize}

\item Randomly explore a first scene where the objects do not overlap and store the experienced sensorimotor transitions $(S_i,\Delta M_q \rightarrow S_j)$.

\item Successively explore new scenes where the objects are randomly translated and overlapped. The environment also a probability of $5\%$ to change randomly in each new scene.

\item During the exploration of new scenes, keep track of the probability of encountering each initial transition $P(S_j|\Delta M_q,S_i)$.

\item Store those probabilities in a matrix $T$ where rows and columns respectively correspond to sensory states $S_i$ and $S_j$ (a single $\Delta M_k$ is associated with each sensory pair and is thus ignored at this step).

\item Cluster the sensory states $S_i$ into subgraphs $Y_a$ based on the statistical co-occurrence of their transitions using spectral clustering on matrix $T$.

\end{itemize}
In order to reduce the computational cost of the algorithm, only ``salient" sensory states $S_i$ are actually stored during the initial exploration (step 1). An arbitrary criterion (contrast in the $3 \times 3$ sensory input) is used to categorize as salient approximately $10\%$ of all sensory inputs the agent can experience\footnote{Any random criterion could be considered here as the only goal of the pre-processing is to reduce memory usage and not to carry information about the actual sensory input.}. This pre-processing is analogous to the initial k-means clustering of the previous simulation.

The algorithm's outcome is a classification of sensory states $S_i$ into subgraphs. Adding back the motor command $\Delta M_q$ that was put aside during the data processing, each subgraph is made of sensory states interconnected with highly predictable transitions, as illustrated in Fig.~\ref{fig:discovering objects}.

The results of the algorithm are presented in Fig.~\ref{fig:discovering objects}. The spectral clustering generates $3$ highly predictable subgraphs and a weakly predictable one.
They respectively correspond to the three objects the agent interacted with and to the environment of the initial scene which became very improbable due to its random changes during exploration.
When recognizing a sensory inputs as part of an object, the agent can predict which sensory input it would receive by navigating into the corresponding subgraph.
A more complete description of the algorithm and discussion about its benefits and limitations are available to the interested reader in \cite{laflaquiere2015object}.


\subsection{Discovering the visual field}
\label{sec:discovering the visual field}

\begin{figure*}[t]
\includegraphics[width=1\linewidth]{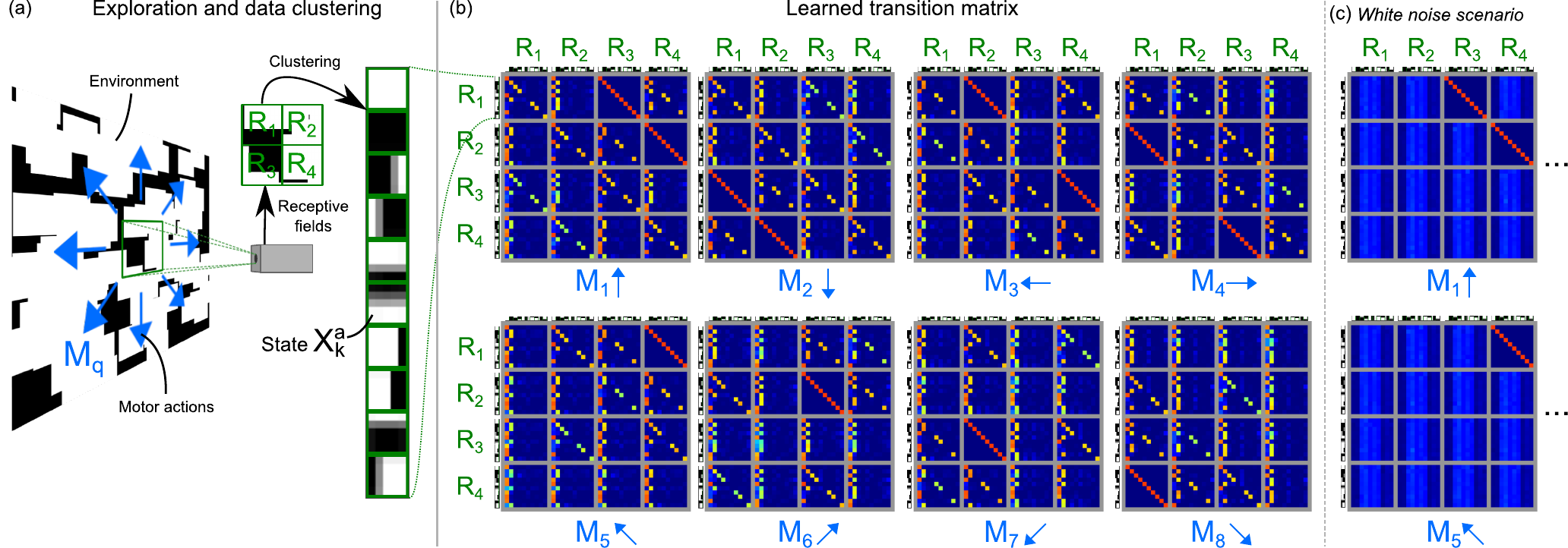}
\caption{
(a) Agent exploring its environment with a simplified retina-like sensor made of 4 receptive fields that can move in 8 directions in a discrete way. The environment is made of random white squares in front of a black background. Data collected during exploration of different environments are clustered using k-means (the same clusters are used for all receptive fields).
(b) The probabilities of transition between clusters of the different receptive fields while the agent moves are estimated based on the collected data. They are stored in a 3D matrix $T$ whose slices are displayed separately.
For each movement of the sensor, a subset of transitions is highly predictable (red diagonals in blocks): they correspond to the shifting of the visual scene on the retina. Other significantly predictable transitions appear in the matrix: they capture regularities in the explored environments.
(d) Those additional predictive transitions disappear when the agent explores structure-less environments (white noise scenes), which highlights the specific highly predictable transitions related to the visual field (red diagonals in block).
}
\label{fig:discovering visual field}
\end{figure*}

The sensorimotor contingencies approach covers more than the perception of environmental properties, as was the case in the two previous simulations. It also proposes to ground the perception of the agent's own body and ability to interact with the world. In this last illustration, we address the problem of \textit{perceiving} (in the sense of \textit{mastering}) the visual field associated with a visual sensor.

The simulation has been inspired by the human visual apparatus in which the retina is covered by multiple receptive fields. They encode the information for small regions of the visual scene \cite{Croner19957}. 
It has been shown that the brain learns the association between the sensory excitation of those different regions as a function of its own saccade commands \cite{herwig2014predicting}. As an example, the brain has to learn that a given sensory input  (e.g. a corner) in its central receptive field corresponds to an equivalent sensory input  (e.g. a potentially differently encoded corner) in its top receptive field when it performs a downward saccade.
All the statistically consistent associations that can be established between the sensory inputs of different receptive fields can be discovered while exploring visual scenes. They define all the predictable transformations that can be generated in the sensory input by saccading and which correspond to the image projection being shifted on the retina. In other words, they constitute the experience of having a \textit{visual field}.

Our objective is for a naive agent equipped with a camera-like sensor to discover those predictable transitions. In order to illustrate this approach, we consider the agent-environment system presented in Fig.~\ref{fig:discovering visual field}. The agent is equipped with a camera and two motors. For the sake of simplicity, the camera spans only a narrow part of the environment ($10 \times 10$ pixels). The captured image is encoded by $4$ adjacent receptive fields of size $5\times 5$ pixels. The sensory input $S_i$ in a receptive field $a$ is denoted $S^a_i$. The two motors allow the camera to translate in the plane. Eight motor commands $\Delta M_q$ are considered in the simulation such that some receptive fields overlap before and after a saccade (as illustrated in Fig.~\ref{fig:discovering visual field}).
The agent can explore visual scenes made of white squares of variable sizes randomly distributed on a black background. Once again those scenes are kept simple to facilitate the results analysis but the approach could naturally apply to more complex environments.

The algorithm we propose to process the agent's sensorimotor experience is as follows:
\begin{itemize}
\item Explore successive environments with random saccades $\Delta M_q$ (motor babbling) to collect sensory experiences $S^a_i$ and 
sensorimotor transitions $(S^a_i, \Delta M_q \rightarrow S^b_j)$.

\item Cluster the sensory data of each receptive field $S^a_i$ into $K=10$ clusters $X^a_k$ using \textit{k}-means in order to discretize the sensory space. Since, statistically, all receptive fields receive the same sensory inputs, 
we fix $X^b_k=X^1_k, \forall \{b,k\}$.

\item Evaluate the probability of transitions $P(X^b_l|\Delta M_q, X^a_k)$ based on the experienced transitions and store them in a 3D matrix $T$ where the first dimension corresponds to a starting state $X^a_k$, the second to an ending state $X^b_l$, and the third to the motor command $\Delta M_q$.

\end{itemize}

Similarly to previous simulations, an additional step could be considered to identify subgraphs in matrix $T$. The outcome would be subgraphs describing how clusters $X^a_k$ are related through saccadic motor commands. However, all interesting results are already displayed in the matrix $T$ and this step is thus skipped.

The simulation results are presented in Fig.~\ref{fig:discovering visual field}.
For each motor command $\Delta M_q$, some sensorimotor transitions have a high probability $P(X^b_l|\Delta M_q, X^a_k)$. They appear as red diagonals in blocks of the matrix $T$ and capture the fact that the sensory input is shifted between the two corresponding receptive fields during the saccade $\Delta M_q$.
Having estimated this predictive model, the agent can thus predict how sensory inputs "move" in the visual field when it performs saccades.

Moreover, other significantly predictable transitions can be observed in $T$. They are induced by structure in the environment (and indirectly in the clusters $X^a_k$). For instance, a state $X^a_k$ corresponding to a vertical edge will predict itself with a high probability for motor commands corresponding to upward or downward saccades. As a reminder of a preceding discussion in \ref{sec:2}, such an invariance of the sensory input to a motor command is what partially defines the interaction with a line.
As can be seen in Fig.~\ref{fig:discovering visual field}, those additional predictable transitions disappear when the agent explores random scenes (white noise). No environment-related structure is then extracted and only transitions related to the structure of the visual field appear with high probability.

\section{CONCLUSION}  
\label{sec:ccl}

This paper presented our developmental approach of sensorimotor contingencies. Its objective has been to address the questions \textit{what is perception} and \textit{how can it be acquired}. The framework we develop stands at the intersection of sensorimotor contingencies theory, developmental robotics and predictive modeling. As discussed in section \ref{sec:2}, SMCT offers a promising approach to define a grounded perceptive ontology but does not address the question of its acquisition. Taking inspiration from animal and human development, we claim that sensorimotor contingencies should be learned incrementally by exploring the world and capturing internal models that allow the robot to predict the potential outcome of its actions.

When considering naive agents without a priori knowledge about the world, many difficulties have to be addressed. 
Based on its sensorimotor experiences, the robot has to identify contingencies, the correct sensorimotor subspace in which they are defined, and when they are relevant to model the data.
Moreover, each modeled contingency extends the already existing sensorimotor space and potentially generates new higher-level ones, leading to a hierarchical structure the agent may discover. 

Some of those difficulties have been addressed in Sec.~\ref{sec:preliminary applications}. Simple simulations have been designed to illustrate the discovery of specific contingencies. Although slightly different algorithms have been proposed to solve each of those problems, they all stem from identical considerations: contingencies imply structure in sensorimotor transformations, and can be captured as such by predictive models.

Our current effort and main objective is to formalize a single and generic learning framework which would
account for the acquisition of any contingency. We aim to
unify the three algorithms presented in this paper 
and address additional problems such as ambiguous experiences, hierarchical structures, and compact encoding of contingencies.
We always have in mind keeping our approach compatible with a developmental approach, even if the developmental setting was presented as contextual, and did not play a key role in our experiments. 
In particular, the agents in our experiments used a purely random policy to explore the space of sensorimotor experiences. This was sufficient for our purposes, as the focus of this work lies more on the question of what representational format can be used to capture sensorimotor contingencies. Mechanisms driving exploration in a more efficient way, such as intrinsic motivation~\cite{oudeyer2007intrinsic}, will surely be necessary to adopt our approach in larger scale scenarios. In future work we therefore intend to incorporate such mechanisms in our framework, along with an incremental building of hierarchical representations of contingencies. We believe that these two aspects will go hand in hand: by discovering and learning sensorimotor contingencies, an agent acquires a more abstract re-description of the raw sensorimotor signals. This opens up the possibility for the agent to also explore the contingencies themselves, which in turn shapes the low-level exploration behavior of the agent, and will hopefully lead to a developmental learning of increasingly abstract concepts.

\addtolength{\textheight}{-12cm}   

\section*{ACKNOWLEDGMENT}

We would like to thank Ahmed Faraz Khan for his help and his contribution to the presented experiments.
 
\bibliographystyle{unsrt}
\bibliography{bib_short}

\end{document}